\documentclass[runningheads]{llncs}

\usepackage{eccv}

\usepackage{eccvabbrv}

\usepackage{graphicx}
\usepackage{booktabs}
\usepackage{color}
\usepackage{amsfonts,amssymb}
\usepackage{multirow, multicol}
\usepackage{mathrsfs}
\usepackage{wrapfig}
\usepackage{threeparttable}
\usepackage{makecell}
\usepackage{arydshln}
\usepackage{bbm}
\usepackage{amsmath}
\usepackage[export]{adjustbox}

\usepackage[accsupp]{axessibility}  

\usepackage{hyperref}

\usepackage{orcidlink}

\begin{document}

\title{Activating Wider Areas in Image Super-Resolution} 

\titlerunning{Activating Wider Areas in Image Super-Resolution}

\author{Cheng Cheng\inst{1} \and
Hang Wang\inst{2} \and
Hongbin Sun\inst{1}$^{,*}$}
\renewcommand{\thefootnote}{\fnsymbol{footnote}}
\footnotetext[1]{Corresponding Author.}
\renewcommand{\thefootnote}{\arabic{footnote}}

\authorrunning{Cheng. et al.}

\institute{College of Artificial Intelligence, Xi'an JiaoTong University \and
School of Microelectronics, Xi'an JiaoTong University \\
\email{cheng2016@stu.xjtu.edu.cn}}

\maketitle

\begin{abstract}

The prevalence of convolution neural networks (CNNs) and vision transformers (ViTs) has markedly revolutionized the area of single-image super-resolution (SISR).
To further boost the SR performances, several techniques, such as residual learning and attention mechanism, are introduced, which can be largely attributed to \emph{a wider range of activated area}, that is, the input pixels that strongly influence the SR results.
However, the possibility of further improving SR performance through another versatile vision backbone remains an unresolved challenge.
To address this issue, in this paper, we unleash the representation potential of the modern state space model, i.e., Vision Mamba (Vim), in the context of SISR.
Specifically, we present three recipes for better utilization of Vim-based models: 1) Integration into a MetaFormer-style block; 2) Pre-training on a larger and broader dataset; 3) Employing complementary attention mechanism, upon which we introduce the MMA.
The resulting network MMA is capable of finding the most relevant and representative input pixels to reconstruct the corresponding high-resolution images. 
Comprehensive experimental analysis reveals that MMA not only achieves competitive or even superior performance compared to state-of-the-art SISR methods but also maintains relatively low memory and computational overheads (e.g., +0.5 dB PSNR elevation on Manga109 dataset with 19.8 M parameters at the scale of $\times2$). 
Furthermore, MMA proves its versatility in lightweight SR applications. 
Through this work, we aim to illuminate the potential applications of state space models in the broader realm of image processing rather than SISR, encouraging further exploration in this innovative direction.
\end{abstract}


\section{Introduction}
\label{sec:intro}

Single image super-resolution (SISR) has long been an attractive task in the area of computer vision, which aims to restore high-resolution images with vivid details and textures given their low-resolution counterpart.
SISR is a fundamental problem and has numerous applications in many image analysis tasks including surveillance and satellite image.
Over recent years, the prevalence of deep learning-based methods~\cite{dong2014learning, wang2018esrgan, gu2019blind, ledig2017photo, bell2019blind, lim2017enhanced, shi2016real, zhang2018image} has significantly advanced this domain and well tackled several related sub-problems, e.g., efficient image SR~\cite{ahn2018fast, hui2018fast, hui2019lightweight} and blind image SR~\cite{gu2019blind, bell2019blind}.

\begin{figure}[ht]
  \centering
  \includegraphics[width=0.95\textwidth]{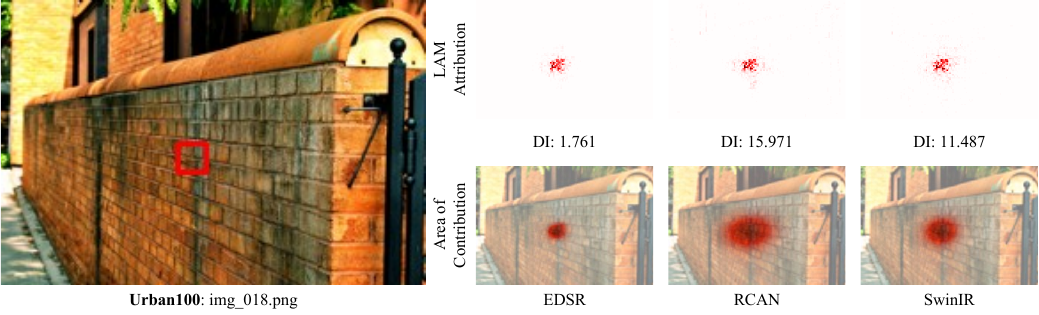}
  \caption{LAM comparison of three representative SR methods. The patches for interpretation are marked with \textcolor{red}{red} boxes in the original images. LAM emphasizes the pixel range engaged in the SR result reconstruction, quantified by the diffusion index (DI). A higher DI indicates that more pixels are involved. Best viewed by zooming.}
  \label{fig:lam}
\end{figure}

From the perspective of the network interpretation~\cite{luo2016understanding, gu2021interpreting}, better SR performances can be attained by expanding effective receptive fields (ERFs)~\cite{luo2016understanding} or the limited local attributed maps (LAMs~\cite{gu2021interpreting}).
Accordingly, existing recipes for boosting SR performances can be summarized as follows.
1. \emph{Stacking multiple layers/blocks.} By stacking multiple layers, convolution neural networks (CNNs) with residual connections~\cite{he2016deep}, e.g., \textbf{EDSR}~\cite{lim2017enhanced}, have shown promising super-resolved performance,.
Besides, these methods pose rather positive scaling properties, that is, deeper and wider networks lead to better performances.
2. \emph{Applying attention mechanism.} To better utilize more neurons for restoring high-resolution images, several attention mechanisms, such as channel attention~\cite{zhang2018image} and pixel attention~\cite{zhao2020efficient}, are introduced to SISR, the most representative one is \textbf{RCAN}.
For instance, channel attention calibrates channel-wise weights to extract more precise and accurate image representation.
Pixel attention performs the attention mechanism on each element of the feature representations, which can be viewed as a more general form of attention operation, thus leading to a better performance.
3. \emph{Utilizing Vision Transformers.} Enlightened by the success of vision transformers (ViTs) in image classification~\cite{dosovitskiy2020image}, \textbf{SwinIR}~\cite{liang2021swinir} showcases that content-based interactions between image content, i.e., spatially varying convolution~\cite{balduzzi2017shattered}, and long-range dependency modeling play a crucial role in SISR.
Despite the aforementioned success, existing SR networks still fail to live up to expectations.
There are still some drawbacks to current techniques.
1) Simply stacking CNN-blocks can hardly obtain better improvements~\cite{zhang2018image} due to training difficulty~\cite{balduzzi2017shattered} and Gaussian property of receptive field~\cite{luo2016understanding}.
2) Utilizing a single attention mechanism may not be the optimal choice, the SR performances may witness further improvements by employing a complementary attention mechanism~\cite{chen2023activating}.
3) To balance efficiency and effectiveness, the self-attention mechanism of Swin-transformer block~\cite{liu2021swin} is performed in ``windows'', which still leaves the global interaction problem unresolved, which is regarded as the key factor for better performances w.r.t computer vision tasks~\cite{gu2021interpreting, wang2018non, liu2022convnet, ding2022scaling}.
Furthermore, as shown in \cref{fig:lam}, though employed with effective techniques, existing networks still retain restricted areas of activated input pixels, resulting in sub-optimal results and facing performance bottlenecks.
Consequently, it is essential to \emph{build up a novel SR network with larger areas of activated input pixels for superior performances}.

To this end, we resort to an up-to-date bidirectional state space model (SSM), i.e., Vision Mamba~\cite{zhu2024vision} (Vim) to form the proposed SR network, MMA (abbreviation for \textbf{M}eet \textbf{M}ore \textbf{A}reas).
To better unleash the representation potential of Vim, we identify the following three key points.
1) Compared to directly replacing CNN blocks or transformer blocks with SSMs as~\cite{ma2024u, ruan2024vm, wang2024mamba, yang2024vivim}, incorporating Vim into a MetaFormer-style~\cite{yu2022metaformer} block can act as a more suitable choice since the general architecture of the transformers is more essential to the resulting performance.
2) Similar to transformer-based counterparts, pre-training Vim-based models on a larger and broader dataset (e.g., ImageNet~\cite{deng2009imagenet}) is of great importance to enhance the model's representation capability.
3) Despite the inherent ability to model global information, applying complementary attention mechanisms together with Vim can further enlarge the overall activated areas, resulting in significant performance elevations.
Compared to state-of-the-art SR networks, either CNN-based (RCAN-it~\cite{lin2022revisiting}) or transformer-based (EDT~\cite{li2021efficient}), MMA demonstrates superior performances and retains relatively low computational overhead (e.g., +0.5 dB PSNR improvement on the Manga109 dataset with 19.8 M parameters at the scale of $\times2$).
The contributions of MMA are three-fold: 1) showing the potential of Vim-based models in the task of SISR; 2) providing effective and efficient recipes of how to better utilize Vim-based models; 3) setting a new benchmark in common SR datasets, e.g., Set5.
Impartial and comprehensive experiments are conducted to verify the effectiveness of the proposed MMA, the commonly used metrics, PSNR, and SSIM witness improvements of 0.4 dB and 0.33 at most in the scale of $\times2$.
Besides, MMA also demonstrates applicability in terms of lightweight SR~\cite{ahn2018fast, hui2019lightweight, li2020lapar, liang2021swinir}, showing superior SR performances.
We hope this work can shed light on the application of modern SSMs~\cite{gu2021efficiently, nguyen2022s4nd, gupta2022diagonal, gu2023mamba, gu2022parameterization, mehta2022long} not only in the area of image SR but the general image processing.

\section{Related Work}
\label{sec:related}

\subsection{Image Super-Resolution}
Since the pioneer work SRCNN~\cite{dong2014learning}, plenty of deep networks \cite{wang2018esrgan, ledig2017photo, lim2017enhanced, shi2016real, zhang2018image} have been proposed and push the frontier of the field of image SR.
These networks, e.g., EDSR~\cite{lim2017enhanced}, usually contain several CNN blocks with residual connection~\cite{he2016deep} to enlarge the receptive field.
Furthermore, attention mechanism~\cite{zhang2018image} and non-local operation~\cite{wang2018non} activate more neurons in SR networks, leading to more vivid details and high-quality high-frequency restoration.
For example, RCAN~\cite{zhang2018image} utilize channel attention~\cite{hu2018squeeze} to formulate the residual in residual structure, as the result of which bringing training stability for deep residual networks and adaptively rescaling features by interdependencies among channel dimension.
As another fundamental backbone for vision tasks, vision transformers~\cite{dosovitskiy2020image} (ViTs) excel in modeling long-range dependency compared to classic CNNs due to its dynamic weights calculation~\cite{han2021demystifying, zhao2021battle} and larger receptive field~\cite{raghu2021vision}.
In practice, to balance effectiveness and efficiency, self-attention operation is performed in ``windows'', i.e., window-based attention.
By employing Swin transformer blocks~\cite{liu2021swin} and residual connection, SwinIR~\cite{liang2021swinir} achieves superior super-resolved results with fewer parameters compared to its CNN counterparts.
Nevertheless, the computational complexity of transformers goes quadratically with the number of image patches, limiting the application to high-resolution images.
Besides, window-based attention is still far from global attention because of the rather limited window size, e.g., $6\times6$.
As demonstrated in previous network interpretation works~\cite{gu2021interpreting}, larger activated areas lead to better performance, hinting that \emph{there is still room for improvements in image SR}.

\subsection{Enlarging Activated Areas in Image Super-Resolution}
Numerous studies have been proposed to explore the potential of attention mechanisms for image SR.
For instance, channel attention, introduced by RCAN~\cite{zhang2018image}, makes it practical to train deep high-performance SR networks.
Instead of performing re-weighting in the channel dimension, PAN~\cite{zhao2020efficient} introduces the pixel attention mechanism which can be viewed as the combination of channel attention and spatial attention~\cite{woo2018cbam}, achieving not only impressive results but retaining few parameters.
After entering the Vision Transformer~\cite{dosovitskiy2020image} era, the application of the attention mechanism has witnessed great changes.
To balance effectiveness and efficiency, as the core of transformer blocks, self-attention operations are performed in ``windows''.
In such a case, SwinIR outperforms its CNN counterparts by a notable margin.
Recently, inspired by the success of large kernel convolutions in high-level tasks~\cite{ding2022scaling, guo2023visual}, VapSR~\cite{zhou2022efficient} utilizes such a paradigm in the attention branch.
By integrating with the novel pixel normalization, VapSR reduces considerable parameters while improving the final performance.

\subsection{State Space Models for Visual Tasks}
Since the success of the Structure State-Space Sequence (S4) model~\cite{gu2021efficiently}, a series of representative models have been introduced to attempt to apply the SSM as a novel alternative to CNNs or transformers.
~\cite{islam2022long} uses a 1D S4 model to deal with the long-range temporal dependencies for video classification. 
Furthermore, S4nd~\cite{nguyen2022s4nd} explores the potential of SSMs in modeling multi-dimensional signals, e.g., 2D images and 3D videos.
TranS4mer~\cite{islam2023efficient} combines the strengths of SSMs and the self-attention mechanism, achieving state-of-the-art performance for movie scene detection. 
By integrating content-based reasoning into classic SSMs, Mamba~\cite{gu2023mamba} not only enjoys fast inference but also the linear scaling property in sequence length. 
Besides, U-Mamba~\cite{ma2024u} also introduces a hybrid CNN-SSM architecture for biomedical image segmentation.
Despite successful applications of the aforementioned works, applying SSMs to image SR is still not explored.

\section{Method}
\label{sec:method}
In this section, we first present concise details of the Vim block and then elaborate on the specifics of the proposed SR network, MMA.
Finally, we give different MMA variants for classic image SR and lightweight SR, respectively.

\subsection{Revisiting Vision Mamba}
\label{sec:method-mamba}
The bidirectional SSM, i.e., Vision Mamba~\cite{zhu2024vision} (Vim), shows superior performance in image classification, object detection, and semantic segmentation.
Originating from classic SSM~\cite{kalman1960new}, Vim excels in modeling long-range dependency and benefits from parallel training.
Compared to modern SSMs~\cite{gu2021efficiently, nguyen2022s4nd, gupta2022diagonal, gu2023mamba, gu2022parameterization, mehta2022long}, e.g., S4~\cite{gu2021efficiently}, and Mamba~\cite{gu2023mamba} the two major contributors of Vim are \emph{bidirectional modeling} and \emph{positional awareness}, which can efficiently compress visual representations and let Vim be more robust in dense prediction tasks, which meets the demands of SISR. The mathematical formulations of SSMs are as follows.

Classic SSMs map a 1-D function $x(t)\in \mathbb{R} \mapsto y(t) \in \mathbb{R}$ through a hidden state $h(t)\in \mathbb{R}^{N}$, given $\mathbf{A}\in\mathbb{R}^{\mathrm{N}\times\mathrm{N}}$ as the evolution parameter and $\mathbf{B}\in\mathbb{R}^{\mathrm{N}\times1}$, $\mathbf{C}\in\mathbb{R}^{1\times\mathrm{N}}$ as the projection parameters.
\begin{equation}
\begin{aligned}
    h^{'}(t)&=\mathbf{A}h(t)+\mathbf{B}x(t),\\
    y(t)&=\mathbf{C}h(t).
\end{aligned}
\label{eq: ssm}
\end{equation}
The discrete version of this linear model can be transformed by zero-order hold (ZOH) given a timescale parameter $\boldsymbol{\mathrm{\Delta}}$.
\begin{equation}
\begin{aligned}
    \overline{\mathbf{A}}&=\mathrm{exp}(\boldsymbol{\mathrm{\Delta}} \mathbf{A}),\\
    \overline{\mathbf{B}}&=(\boldsymbol{\mathrm{\Delta}} \mathbf{A})^{-1}(\mathrm{exp}(\boldsymbol{\mathrm{\Delta}} \mathbf{A})-\mathbf{I})\cdot \boldsymbol{\mathrm{\Delta}} \mathbf{B}.
\end{aligned}
\end{equation}
Then the discretized version of \cref{eq: ssm} can be rewritten as below:
\begin{equation}
\begin{aligned}
    h_{t}&=\overline{\mathbf{A}}h_{t-1}+\overline{\mathbf{B}}x_{t},\\
    y_{t}&=\mathbf{C}h_{t}.
\end{aligned}
\label{eq: dis}
\end{equation}
Finally, the output of such discretized SSM is computed via a global convolution.
\begin{equation}
\begin{aligned}
    \overline{\mathbf{K}}&=(\mathbf{C}\overline{\mathbf{B}}, \mathbf{C}\overline{\mathbf{AB}}, ..., \mathbf{C}\overline{\mathbf{A}}^{\mathrm{M}-1}\overline{\mathbf{B}}),\\
    \mathbf{y}&=\mathbf{x}*\overline{\mathbf{K}},
\end{aligned}
\label{eq: dis-out}
\end{equation}
where $\mathbf{M}$ is the length of input sequence $\mathbf{x}$ and $\overline{\mathbf{K}}\in \mathbb{R}^{\mathbf{M}}$ is a structured convolutional kernel.

As for Vim, to process vision task, 2D images $\mathrm{i} \in \mathbb{R}^{\mathrm{H}\times\mathrm{W}\times\mathrm{C}}$ into flatten 1D non-overlapping patches $\mathrm{i}_\mathrm{p} \in \mathbb{R}^{\mathrm{J}\times(\mathrm{P}^2\cdot\mathrm{C})}$, where $\mathrm{C}$ is the number of channels, $\mathrm{J}$ is the number of patches and $\mathrm{P}$ is the size of patches.
Next, Vim linearly project $\mathrm{i}_\mathrm{p}$ to the vector with size $\mathrm{D}$ and add position embeddings $\mathrm{E}_{pos} \in \mathbb{R}^{(\mathrm{J}+1)\times\mathrm{C}}$ to get the token sequence.
Inside the state space block, different from Mamba, Vim processes the projected vector from both \emph{forward} and \emph{backward} directions, and the output of these two operations is fused by a gated mechanism.
For a comprehensive understanding, we refer readers to \cite{zhu2024vision} for details.

\begin{figure}[htb]
  \centering
  \includegraphics[width=0.95\textwidth]{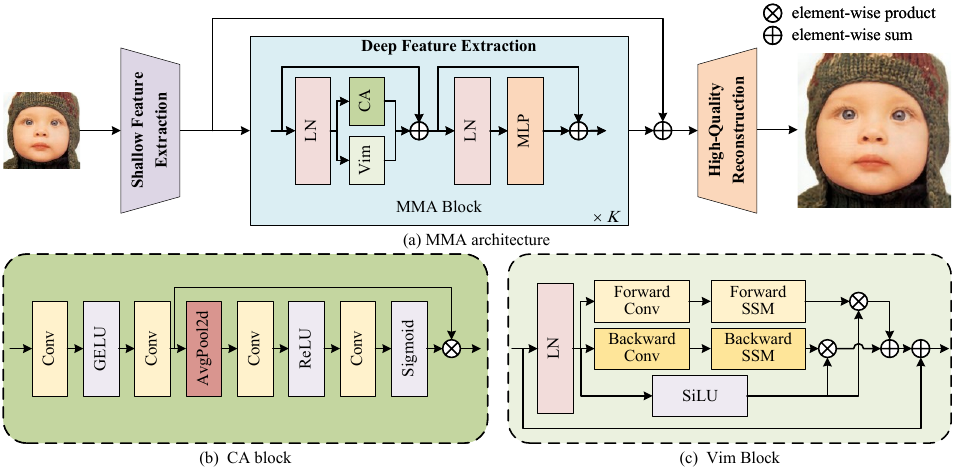}
  \caption{(a) The network architecture of our MMA and the structure of MMA block; (b) The structure of channel attention (CA) block; (c) The structure of Vim block.}
  \label{fig:main}
\end{figure}

\subsection{Elaboration of MMA}

It can be observed from \cref{fig:main} that the proposed MMA consists of three major modules, i.e., shallow feature extraction $H_{SF}$, deep feature extraction $H_{DF}$, and high-quality reconstruction modules $H_{REC}$.

Given a low-resolution image $I_{lr}\in\mathbb{R}^{\mathrm{H}\times\mathrm{W} \times3}$ ($\mathrm{H}$ and $\mathrm{W}$ are the image height and weight), a $3\times3$ convolutional layer is utilized as $H_{SF}(\cdot)$ to extract shallow features $F_{SF}\in\mathbb{R}^{\mathrm{H}\times\mathrm{W} \times \mathrm{C}}$, where $\mathrm{C}$ is the number of features.
\begin{equation}
    F_{SF}=H_{SF}(I_{lr})
\end{equation}
Designing the shallow feature extraction module is a trivial task. 
A simple convolution layer can serve as a good feature extractor at early visual processing for stable optimization and better performances~\cite{xiao2021early}.
Besides, it can be also viewed as a neat way to map the input image to a higher dimensional feature space.

Then the deep feature $F_{DF}$ is extracted by $H_{DF}$, which consists of $K$ MMA blocks.
\begin{equation}
    F_{DF}=H_{DF}(F_{SF})
\end{equation}
Specifically, a single MMA block takes the MetaFormer-style, that is, consists of two Layer Normalization layers~\cite{ba2016layer}, a \emph{token mixer}~\cite{yu2022metaformer} (including a Vim block and a channel attention block in parallel), and an MLP block, as shown in \cref{fig:main} (a).
The reasons are two-fold.
First, despite the powerful capability to model \emph{global interactions} of Vision Mamba~\cite{zhu2024vision}, how to incorporate it into the current basic network architecture still remains unresolved.
Directly replacing CNN-blocks with Mamba like-blocks~\cite{liu2024swin, behrouz2024graph, wang2024semi, wang2024mamba, ruan2024vm} may lead to sub-optimal solutions or face performance bottlenecks.
As demonstrated in~\cite{yu2022metaformer}, as long as a model adopts MetaFormer-style as the general architecture, promising results could be attained.
Therefore, we integrate Vim into the MetaFormer-style block.
Second, to better unleash and utilize the global information extracted by the Vim block, we utilize a channel attention block as the complement.
Despite its simplicity, channel attention can activate more pixels as global information is involved in calculating the channel attention weights.
Besides, it is straightforward to hold the opinion that applying a convolution-based module can help Vim get a better visual representation and ease the training procedure as it does to transformers~\cite{yuan2021incorporating, li2023uniformer, wu2021cvt}.

Finally, the super-resolved image is generated by aggregating both shallow and deep features as follows:
\begin{equation}
    I_{sr}=H_{REC}(F_{SF}+F_{DF}),
\end{equation}
where the pixel-shuffle~\cite{shi2016real} is utilized as the upsampling operation for $H_{REC}$. 
Shallow features mainly contain low-frequencies information while deep features focus on restoring missing high-frequencies information.
Therefore, with a shortcut connection, the proposed MMA can transmit complementary information directly to $H_{REC}$, which will stabilize the training procedure~\cite{liang2021swinir}.

During training, we optimize MMA by minimizing $L_1$ pixel loss
\begin{equation}
    \mathscr{L}=||I_{sr}-I_{hr}||_1,
\end{equation}
where $I_{hr}$ is the corresponding grounding-truth high-resolution image.

\subsection{Pre-training Scheme}
Previous studies have demonstrated that CNN-based high-level vision tasks~\cite{kornblith2019better, mahajan2018exploring, sun2017revisiting} pre-trained on ImageNet~\cite{deng2009imagenet} classification result in notable performance improvements.
Furthermore, a recent study, EDT~\cite{li2021efficient}, further points out that transformer-based low-level vision models, e.g., image SR and denoising, can benefit from pre-training on ImageNet in a multi/single-task setup.
Theoretically, the main reason why pre-training improves the model performance is by introducing different degrees of local information, which can be treated as a kind of inductive bias to the intermediate layers.

As either CNN-based or transformer-based models can be boosted by pre-training on large-scale datasets, it is straightforward to ask the question: \emph{Can Vim-based model also benefit from pre-training}? 
Therefore, different from previous studies on investigating pre-training on either CNN-based or transformer-based models, in this work, we explore the applicability of pre-training w.r.t. Vim-based models.
Through extensive experiments, we can conclude that pre-training is of great importance to Vim-based models.
Training details of this pre-training scheme can be found in \cref{sec:training-details}.

\subsection{MMA Variants}

For the classic image SR task, we set the number of features $C$ to 192 and the number of MMA blocks $K$ to 24, which forms the MMA-B.
For the lightweight SR, we set the number of features $C$ to 48 and the number of MMA blocks $K$ to 8, which forms the MMA-T.
We term MMA-B as MMA for brevity in the following sections.


\section{Experiments}
\label{sec:exps}

\subsection{Experimental Setup}

\subsubsection{Datasets and Metrics}
ImageNet~\cite{deng2009imagenet} is chosen as the pre-training dataset for its broadness and DF2K is chosen as the fine-tuning dataset, which comprises a total of 3,450 LR-HR RGB image pairs, including 800 from DIV2K \cite{agustsson2017ntire} and 2,650 from Flickr2K \cite{lim2017enhanced}. 
For evaluation datasets, we use five widely used benchmark datasets: Set5 \cite{bevilacqua2012low}, Set14 \cite{zeyde2012single}, BSD100 \cite{martin2001database}, Urban100 \cite{huang2015single}, and Manga109 \cite{matsui2017sketch}. 
The widely used metrics, i.e., average peak signal-to-noise ratio (PSNR) and the structural similarity index (SSIM) are used. 
To keep consistency with previous methods \cite{lim2017enhanced, zhang2018image}, we calculate the values on the luminance channel, i.e., the Y channel of the YCbCr color space converted from the RGB channels. 
For the comparison among lightweight SR methods, we also provide the number of \emph{\#Params} of different models.

\subsubsection{Implementation Details}
\label{sec:training-details}
During both pre-training and fine-tuning, we randomly crop patches with the size of $64 \times 64$ from the dataset as input and augment the training data with random horizontal flips and rotations ($90^{\circ}$, $180^{\circ}$, and $270^{\circ}$). 
The LR images are obtained by the bicubic interpolation downsampling operation in MATLAB from corresponding ground-truth images with the scaling factors: $\times 2$, $\times 3$, and $\times 4$. 
The batch size is set to 64 during both pre-training and fine-tuning.
The total training iterations are set to 800K and 250K for pre-training and fine-tuning, respectively. 
The learning rate for pre-training is initialized as $2\times 10^{-4}$ and reduced by half at the milestone [300K,500K,650K,700K,750K] while the learning rate of fine-tuning is initialized as $1\times 10^{-5}$ and reduced by half at the milestone [125K,200K,225K,240K].
To optimize the model, we adopt the $L1$ loss function and the ADAM optimizer \cite{kingma2014adam} with $\beta 1=0.9$  and $\beta 2=0.99$ to train the model. 
The exponential moving average (EMA) decay is set to 0.999 to stabilize the training. 
The $\times 2$, $\times 3$, and $\times 4$ models are all trained from scratch. 
All experiments are conducted with the PyTorch framework \cite{paszke2019pytorch} on four NVIDIA GeForce RTX 3090 GPUs.

\subsection{Visual Comparison on Image SR}
\begin{figure}[p]
  \centering
  \includegraphics[width=1\textwidth]{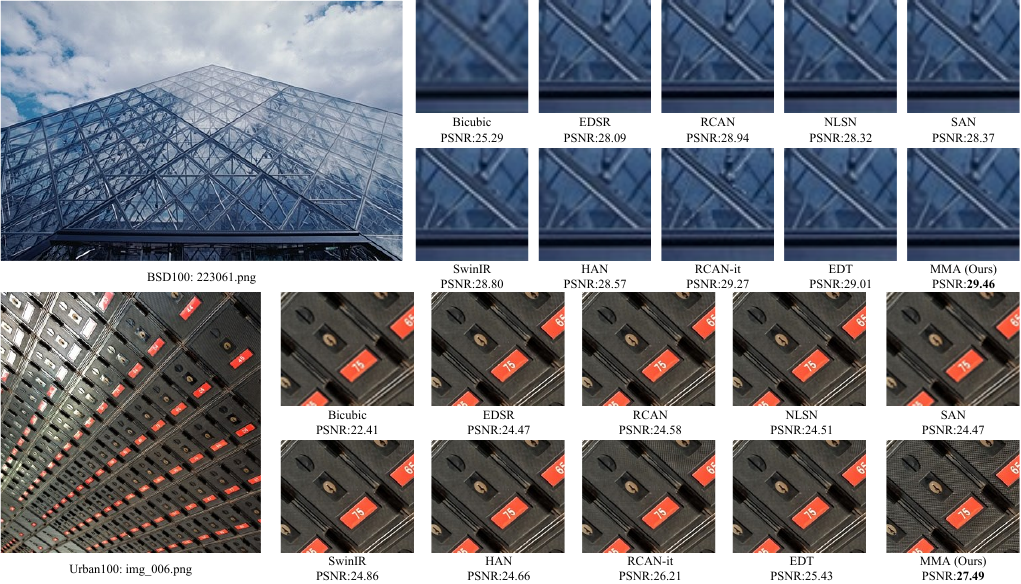}
  \caption{Visual comparison ($\times2$) between MMA and state-of-the-art SISR methods on BSD100 and Urban100 datasets. Best viewed by zooming. The highest PSNR are marked in \textbf{bold}. More visual results are provided in the supplementary material.}
  \label{fig:main-results}
\end{figure}

We provide the visual comparison in \cref{fig:main-results} of MMA and the state-of-the-art methods: EDSR~\cite{lim2017enhanced}, RCAN~\cite{zhang2018image}, SAN~\cite{dai2019second}, NLSN~\cite{mei2021image}, HAN~\cite{niu2020single}, RCAN-it~\cite{lin2022revisiting}, SwinIR~\cite{liang2021swinir}, and EDT~\cite{li2021efficient}. 
As one can see, MMA achieves the best performance among existing SR methods.
Taking the image ``img\_006.png'' in the Urban100 dataset for example, all existing SR methods fail to recover detailed textures of the cabinet while MMA not only successfully restores vivid and authentic textures but also retains sharp edges.
Therefore, it can be concluded that the impressive super-resolved result can not be generated by just considering only area-restricted information, demonstrating the superiority of MMA in modeling long-range dependencies. 
Compared to existing methods, the proposed MMA not only poses leading quantitative results w.r.t. PSNR but also obtains significantly clearer textures than other methods. 

\begin{figure}[p]
  \centering
  \includegraphics[width=1\textwidth]{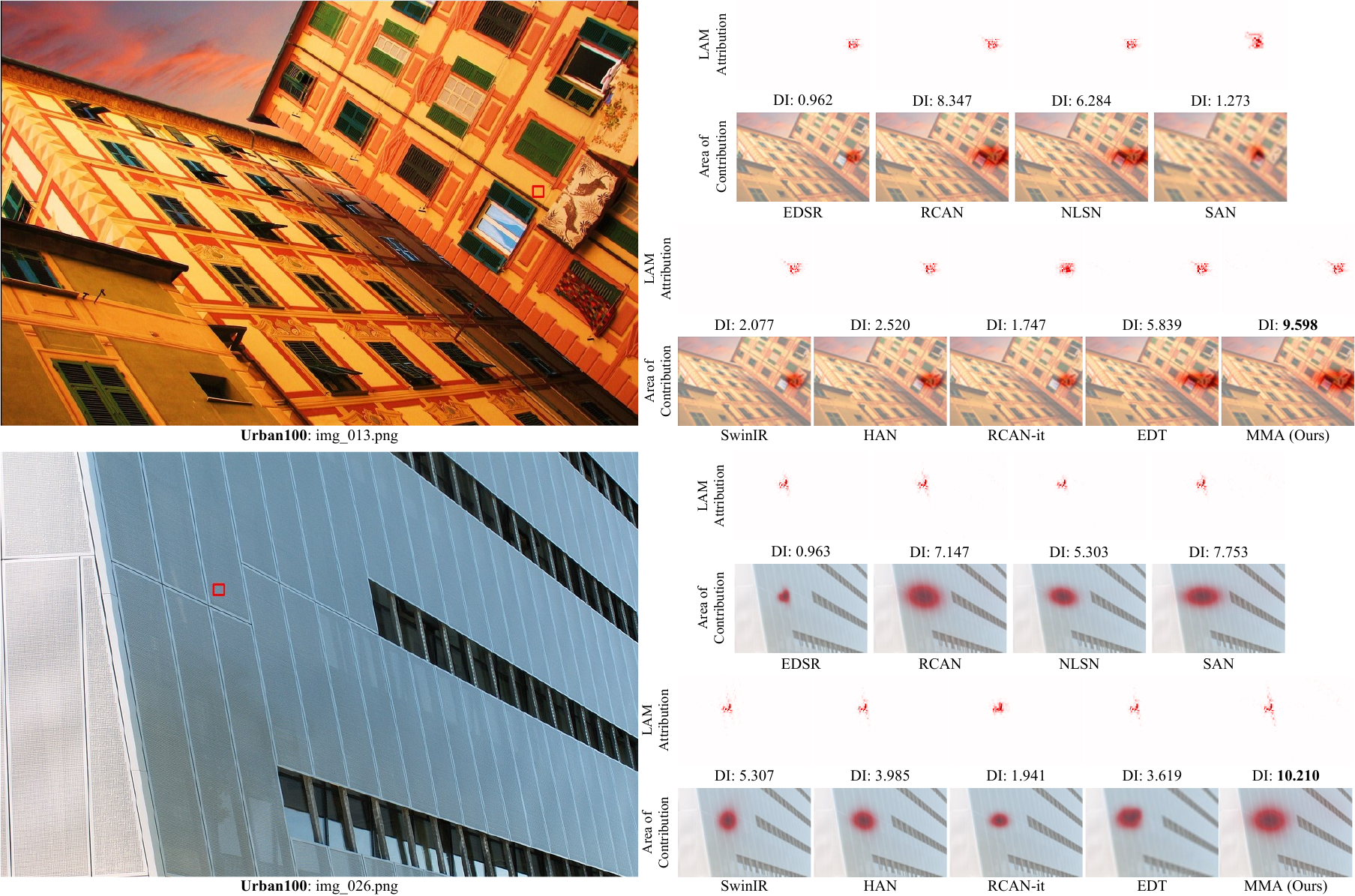}
  \caption{LAM comparison ($\times2$) between MMA and state-of-the-art SISR methods on Urban100 datasets. A higher DI indicates that more pixels are involved. Best viewed by zooming. More LAM results are provided in the supplementary material.}
  \label{fig:main-lam}
\end{figure}

\subsection{Quantitative Results on Image SR}
\label{sect:figures}

\begin{figure}[ht]
  \centering
  \includegraphics[width=0.4\textwidth]{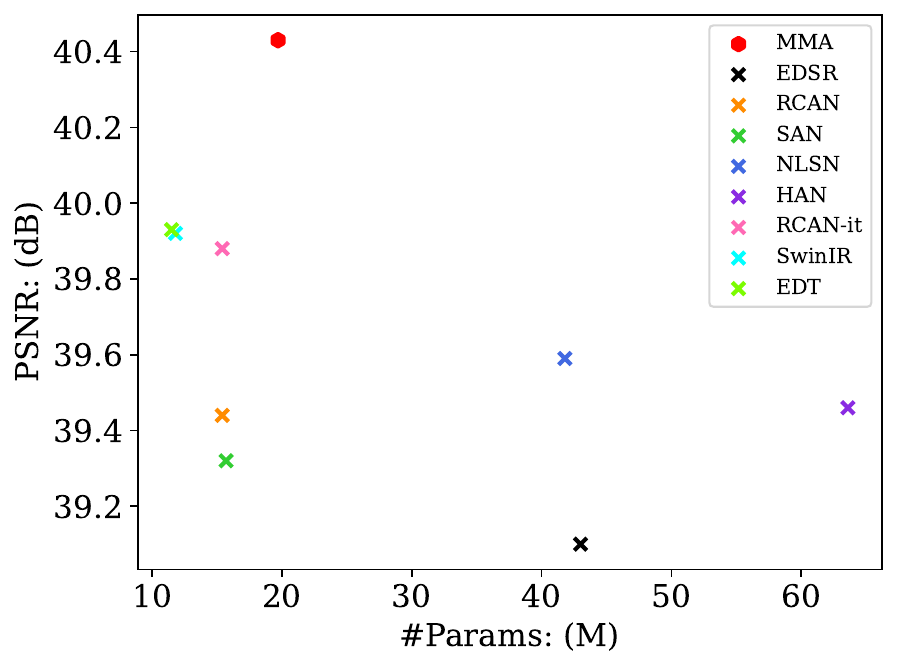}
  \caption{Model complexity comparison ($\times2$). PSNR (dB) on Manga109, \#Params are reported.}
  \label{fig:param}
\end{figure}

We provide a comprehensive quantitative comparison in \cref{tab:main-qualtitative-results}.
It can be observed that the proposed MMA outperforms the other methods significantly on all benchmark datasets. 
Concretely, for all three upsampling scales and all five benchmarks, MMA shows consistent and clear improvements, e.g., MMA outperforms the state-of-the-art method, i.e., EDT, by an elevation of 0.4 dB in terms of PSNR at a scale of $\times2$.
Besides, we also provide the model complexity comparison in \cref{fig:param}.
Compared to existing SR methods, MMA retains a rather modest computation overhead.
All these quantitative results show the effectiveness and efficiency of MMA.

\begin{table*}[ht]
\centering
\caption{Quantitative comparison between MMA and state-of-the-art SISR methods. The best is highlighted in \textcolor{red}{red} and the second is \underline{underlined}.} 
\label{tab:main-qualtitative-results}
\resizebox{0.95\textwidth}{!}{
\begin{threeparttable}
\begin{tabular}{l*{9}{c}}
\bottomrule[2pt]

\multirow{2}{*}{Method} & \multirow{2}{*}{Scale} &   Set5      &   Set14     & BSD100    & Urban100    & Manga109   \\ \cmidrule(r){3-7}
& &   PSNR/SSIM &   PSNR/SSIM & PSNR/SSIM   & PSNR/SSIM & PSNR/SSIM   \\ \hline

Bicubic  & \multirow{10}{*}{$\times 2$} & 33.66 / 0.9299       & 30.24 / 0.8688       & 29.56 / 0.8431       & 26.88 / 0.8403     & 30.80 / 0.9339   \\
EDSR     & & 38.11 / 0.9602       & 33.92 / 0.9195       & 32.32 / 0.9013       & 32.93 / 0.9351     & 39.10 / 0.9773   \\
RCAN     & & 38.27 / 0.9614       & 34.12 / 0.9216       & 32.41 / 0.9027       & 33.34 / 0.9384     & 39.44 / 0.9786   \\
SAN      & & 38.31 / 0.9620       & 34.07 / 0.9213       & 32.42 / 0.9028       & 33.10 / 0.9370     & 39.32 / 0.9792   \\
NLSN     & & 38.34 / 0.9618       & 34.08 / 0.9231       & 32.43 / 0.9027       & 33.42 / 0.9394     & 39.59 / 0.9789   \\
HAN      & & 38.27 / 0.9614       & 34.16 / 0.9217       & 32.41 / 0.9027       & 33.35 / 0.9385     & 39.46 / 0.9785   \\
RCAN-it  & & 38.37 / 0.9620       & 34.49 / 0.9250       & 32.48 / 0.9034       & 33.62 / 0.9410     & 39.88 / 0.9799   \\
SwinIR   & & 38.42 / 0.9623       & 34.46 / 0.9250       & \underline{32.53} / \underline{0.9041}       & \underline{33.81} / \underline{0.9427}     & 39.92 / 0.9797   \\
EDT     & & \underline{38.45} / \underline{0.9624}       & \underline{34.57} / \underline{0.9258}       & 32.52 / \underline{0.9041}       & 33.80 / 0.9425     & \underline{39.93} / \underline{0.9800}   \\ \hdashline
\textbf{MMA} (Ours)     & & \textcolor{red}{38.53} / \textcolor{red}{0.9633}       & \textcolor{red}{34.76} / \textcolor{red}{0.9262}       & \textcolor{red}{32.63} / \textcolor{red}{0.9059}       & \textcolor{red}{34.13} / \textcolor{red}{0.9446}     & \textcolor{red}{40.43} / \textcolor{red}{0.9814}   \\ \midrule[1pt]

Bicubic  & \multirow{10}{*}{$\times 3$} & 33.66 / 0.9299       & 30.24 / 0.8688       & 29.56 / 0.8431       & 26.88 / 0.8403     & 30.80 / 0.9339   \\
EDSR     & & 34.65 / 0.9280       & 30.52 / 0.8462       & 29.25 / 0.8093       & 28.80 / 0.8653     & 34.17 / 0.9476   \\
RCAN     & & 34.75 / 0.9300       & 30.59 / 0.8476       & 29.33 / 0.8112       & 28.93 / 0.8671     & 34.30 / 0.9494   \\
SAN      & & 34.74 / 0.9299       & 30.65 / 0.8482       & 29.32 / 0.8111       & 29.09 / 0.8702     & 34.44 / 0.9499   \\
NLSN     & & 34.85 / 0.9306       & 30.70 / 0.8485       & 29.34 / 0.8117       & 29.25 / 0.8726     & 34.57 / 0.9508   \\
HAN      & & 34.75 / 0.9299       & 30.67 / 0.8483       & 29.32 / 0.8110       & 29.10 / 0.8705     & 34.48 / 0.9500   \\
RCAN-it  & & 34.86 / 0.9308       & 30.76 / 0.8505       & 29.39 / 0.8125       & 29.38 / 0.8755     & 34.92 / 0.9520   \\
SwinIR   & & \underline{34.97} / \underline{0.9318}       & \underline{30.93} / \underline{0.8534}       & \underline{29.46} / \underline{0.8145}       & \underline{29.75} / \underline{0.8826}     & 35.12 / \underline{0.9537}   \\
EDT     & & \underline{34.97} / 0.9316       & 30.89 / 0.8527       & 29.44 / 0.8142       & 29.72 / 0.8814     & \underline{35.13} / 0.9534   \\ \hdashline
\textbf{MMA} (Ours)     & & \textcolor{red}{35.03} / \textcolor{red}{0.9324}       & \textcolor{red}{31.06} / \textcolor{red}{0.8540}       & \textcolor{red}{29.52} / \textcolor{red}{0.8173}       & \textcolor{red}{29.93} / \textcolor{red}{0.8829}     & \textcolor{red}{35.41} / \textcolor{red}{0.9546}   \\ \midrule[1pt]

Bicubic  & \multirow{10}{*}{$\times 4$} & 33.66 / 0.9299       & 30.24 / 0.8688       & 29.56 / 0.8431       & 26.88 / 0.8403     & 30.80 / 0.9339   \\
EDSR     & & 32.46 / 0.8968       & 28.80 / 0.7876       & 27.71 / 0.7420       & 26.64 / 0.8033     & 31.02 / 0.9148   \\
RCAN     & & 32.63 / 0.9002       & 28.87 / 0.7889       & 27.77 / 0.7436       & 26.82 / 0.8087     & 31.22 / 0.9173   \\
SAN      & & 32.64 / 0.9003       & 28.92 / 0.7888       & 27.78 / 0.7436       & 26.79 / 0.8068     & 31.18 / 0.9169   \\
NLSN     & & 32.59 / 0.9000       & 28.87 / 0.7891       & 27.78 / 0.7444       & 26.96 / 0.8109     & 31.27 / 0.9184   \\
HAN      & & 32.64 / 0.9002       & 28.90 / 0.7890       & 27.80 / 0.7442       & 26.85 / 0.8094     & 31.42 / 0.9177   \\
RCAN-it  & & 32.69 / 0.9007       & 28.99 / 0.7922       & 27.87 / 0.7459       & 27.16 / 0.8168     & 31.78 / 0.9217   \\
SwinIR   & & \underline{32.92} / \underline{0.9044}       & \underline{29.09} / \underline{0.7950}       & \underline{27.92} / \underline{0.7489}       & 27.45 / \underline{0.8254}     & 32.03 / \underline{0.9260}   \\
EDT     & & 32.82 / 0.9031       & \underline{29.09} / 0.7939       & 27.91 / 0.7483       & \underline{27.46} / 0.8246     & \underline{32.05} / 0.9254   \\ \hdashline
\textbf{MMA} (Ours)     & & \textcolor{red}{32.96} / \textcolor{red}{0.9645}       & \textcolor{red}{29.12} / \textcolor{red}{0.7966}       & \textcolor{red}{28.01} / \textcolor{red}{0.7524}       & \textcolor{red}{27.63} / \textcolor{red}{0.8274}     & \textcolor{red}{32.49} / \textcolor{red}{0.9282}   \\ \bottomrule[2pt] 

\end{tabular}
\end{threeparttable}
}
\end{table*}

\subsection{LAM Results among Image SR Methods}

\textbf{LAM interpretations}
Local attribution map (LAM) is an interpretation method for SISR, which inherits the integral gradient method~\cite{sundararajan2017axiomatic}.
The \emph{i}th dimension of the LAM is defined as follows:
\begin{equation}
\mathrm{LAM}_{F,D}(\gamma)_{i} := \int_{0}^{1}\frac{\partial D(F(\gamma(\alpha))}{\partial \gamma(\alpha)_{i}} \times \frac{\partial \gamma(\alpha)_{i}}{\partial \alpha} d\alpha ,
\end{equation}
where $F: \mathbb{R}^{h\times w} \mapsto \mathbb{R}^{sh\times sw}$ stands for an SR network with the upscale factor $s$; $D: \mathbb{R}^{l\times l} \mapsto \mathbb{R}$ represents the detector, e.g., gradient detector; $\gamma(\alpha): [0,1] \mapsto \mathbb{R}^{h \times w}$ is a smooth path function; $\gamma$ denotes the degree of Gaussian noise.
Intuitively, LAM finds the input pixels that strongly influence the SR results, which can serve as a good indicator for interpretation. 

We provide the LAM comparison in \cref{fig:main-lam}.
In comparison to existing SR methods, MMA demonstrates superior performance, evidenced by the highest DI and the largest activated area.
This advancement is attributed to the exemplary long-range representational capabilities of Vim coupled with the synergistic integration of a complementary attention mechanism.
As mentioned in prior discussions, the extent of activated regions is highly correlated to the efficacy of SR performance. 
Consequently, the empirical results unequivocally establish that MMA outperforms existing methods by significant margins, setting a new benchmark in the domain of image SR.

\subsection{Ablation studies}
\begin{figure}[ht]
  \centering
  \includegraphics[width=0.82\textwidth]{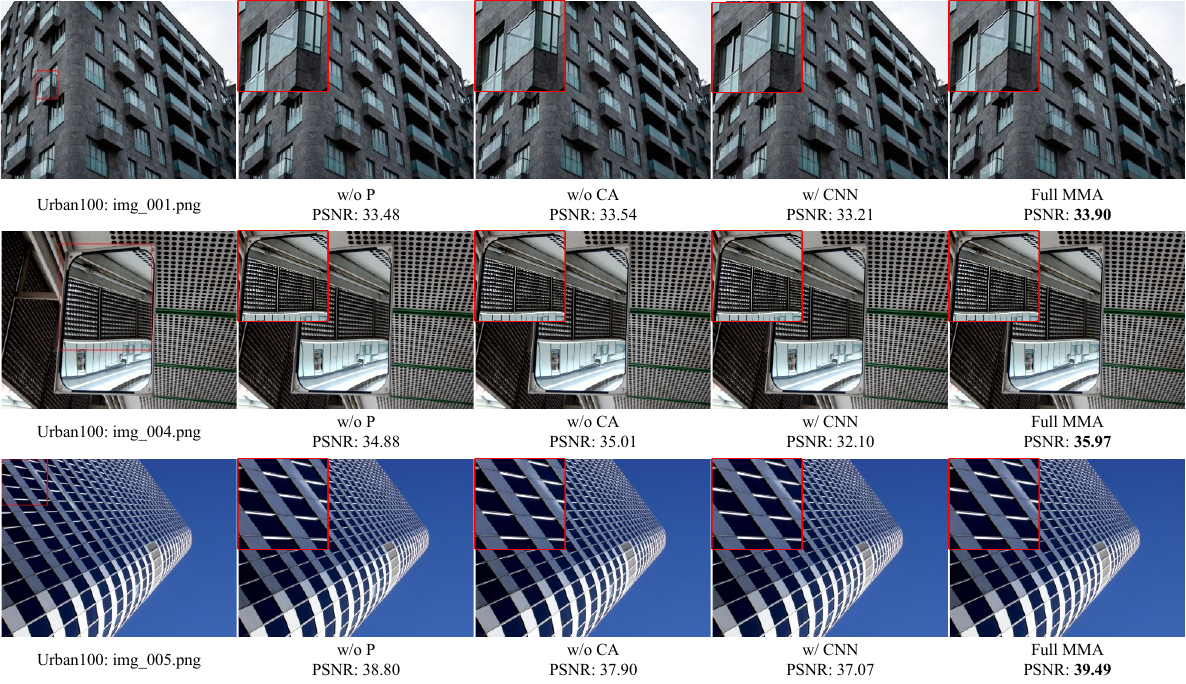}
  \caption{Visual comparison ($\times2$) of several ablation studies on Urban100 datasets.}
  \label{fig:ablation}
\end{figure}

\begin{figure}[ht]
  \centering
  \includegraphics[width=0.82\textwidth]{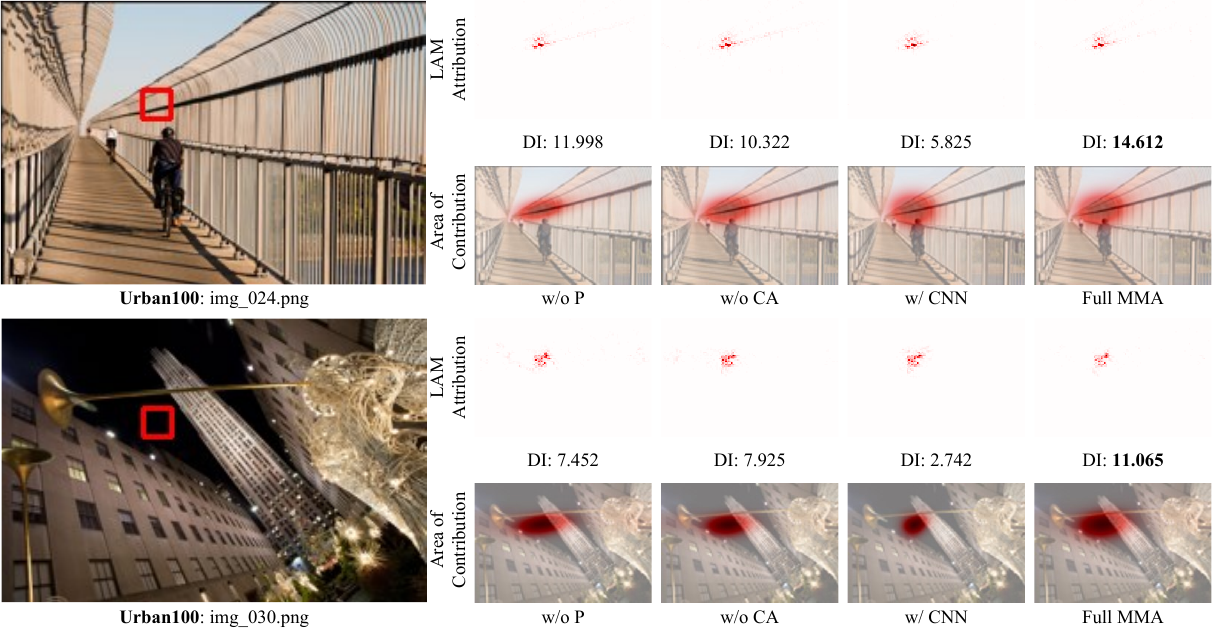}
  \caption{LAM comparison of several ablation studies. The patches for interpretation are marked with \textcolor{red}{red} boxes in the original images. Best viewed by zooming.}
  \label{fig:lam-ablation}
\end{figure}

To further verify the effectiveness of the key designs w.r.t MMA, several ablation studies are conducted with the same training details as the main experiments in \cref{sec:training-details}.
Besides, we also provide the visual comparison in \cref{fig:ablation} and the LAM interpretations of these ablation studies in \cref{fig:lam-ablation}.
Details about these ablation studies are as follows.

\begin{table*}[htb]
\centering
\caption{Quantitative comparison ($\times2$) of several ablation studies. The best is highlighted in \textcolor{red}{red} and the second is \underline{underlined}.} 
\label{tab:ablation}
\resizebox{0.9\textwidth}{!}{
\begin{threeparttable}
\begin{tabular}{l*{9}{c}}
\bottomrule[2pt]

\multirow{2}{*}{Method} &   Set5      &   Set14     & BSD100    & Urban100    & Manga109   \\ \cmidrule(r){2-10}
&   PSNR/SSIM &   PSNR/SSIM & PSNR/SSIM   & PSNR/SSIM & PSNR/SSIM   \\ \midrule

w/o P   & \underline{38.43} / \underline{0.9628}       & \underline{34.50} / \underline{0.9246}       & 32.50 / 0.9046       & \underline{33.65} / \underline{0.9411}     & \underline{39.94} / 0.9801   \\
w/o CA     & 38.37 / 0.9627       & 34.22 / 0.9232       & \underline{32.51} / \underline{0.9048}       & 33.48 / 0.9403     & 39.54 / \underline{0.9804}   \\
w/ CNN     & 38.23 / 0.9620       & 33.99 / 0.9215       & 32.39 / 0.9033       & 33.00 / 0.9361     & 39.11 / 0.9793   \\
\textbf{Full MMA}     & \textcolor{red}{38.53} / \textcolor{red}{0.9633}       & \textcolor{red}{34.76} / \textcolor{red}{0.9262}       & \textcolor{red}{32.63} / \textcolor{red}{0.9059}       & \textcolor{red}{34.13} / \textcolor{red}{0.9446}     & \textcolor{red}{40.43} / \textcolor{red}{0.9814}   \\ \bottomrule[2pt] 

\end{tabular}
\end{threeparttable}
}
\end{table*}

\begin{itemize}
    \item \textbf{Pre-training} 
    Compared to classic image SR methods, a major difference is that MMA adopts the ImageNet pre-training scheme.
    As suggested in previous studies~\cite{gu2023mamba, zhu2024vision}, SSM-based models are data-hungry, therefore pre-training on a larger and more generic dataset, e.g., ImageNet~\cite{deng2009imagenet}, can better unleash the representative potential of SSMs.
    To quantify the impact of this pre-training scheme, we conduct an experiment that directly trains MMA on the DF2K dataset. 
    Other training details are the same as in \cref{sec:training-details}.
    For simplicity, we denote this ablation study as ``$\mathrm{w/o}$ $\mathrm{P}$'' in \cref{tab:ablation} and \cref{fig:ablation}.
    \item \textbf{Channel Attention}
    As pointed out in the aforementioned session, the channel attention mechanism serves as a complementary module to utilize more pixels for reconstruction.
    Intuitively, applying the channel attention mechanism enlarges the activated areas, resulting in more accurate and more precise detail restoration.
    For this reason, it is of great importance to quantify the degree of performance improvements that the channel attention mechanism offers.
    For simplicity, we denote this ablation study as ``$\mathrm{w/o}$ $\mathrm{CA}$'' in \cref{tab:ablation} and \cref{fig:ablation}.
    \item \textbf{Other Type of Token Mixer}
    From previous experiments, it can be observed that the Vim block outperforms classic CNN blocks or transformer blocks in terms of \emph{global} attention mechanism. 
    To quantify the effectiveness of the Vim block in modeling long-range dependency, we conduct another experiment that replaces the Vim block with a classic residual CNN block~\cite{lim2017enhanced} which contains two convolutional layers and one activation layer (ReLU) with a shortcut connection. 
    It should be noted that a MetaFormer-style block resembles a classic vision transformer block, e.g., the Swin-Transformer block. 
    Therefore, the superiority of the Vim block can be easily summarized from previous experiments.
    For simplicity, we denote this ablation study as ``$\mathrm{w/}$ $\mathrm{CNN}$'' in \cref{tab:ablation} and \cref{fig:ablation}.
\end{itemize}

\cref{tab:ablation} reports the quantitative results and \cref{fig:ablation} shows the visual comparison of these ablation studies. 
It can be observed that the pre-training scheme, the complementary channel attention mechanism, and the Vim block all contribute greatly to the MMA.
Removing either of these three designs leads to notable performance drops, especially the Vim block.
Directly replacing the Vim block with a classic CNN block results in 1.6 dB drops w.r.t. PSNR on the Manga109 dataset, hinting that the excellent long-range modeling capability is of great importance to the proposed MMA.

Furthermore, \cref{fig:lam-ablation} demonstrates the LAM interpretations of these ablation studies.
It can be observed that removing channel attention layers and replacing Vim blocks lead to a more restricted activated area compared to the full MMA.
This phenomenon further verifies the effectiveness and superiority of the architectural design of the proposed MMA.

\subsection{Quantitative Results on Lightweight Image SR}

We also provide a comparison of MMA-T with state-of-the-art lightweight image SR methods: CARN~\cite{ahn2018fast}, IMDN~\cite{hui2019lightweight}, LAPAR-A~\cite{li2020lapar} and SwinIR-T~\cite{liang2021swinir}. 
In addition to PSNR and SSIM, we also report the total number of parameters, \emph{\#Params}, to compare the model size of different models. 
As shown in \cref{tab:lightweight-sr}, MMA-T poses competitive or even superior quantitative results over existing methods with similar \emph{\#Params}.
In terms of PSNR, a notable 0.25dB improvement is witnessed on the Manga109 dataset in the scale of $\times2$.  
This further demonstrates the efficiency of the macro-design (MetaFormer-style and parallel token mixer) of the MMA-T architecture.

\begin{table*}[ht]
\centering
\caption{Quantitative comparison between MMA and state-of-the-art lightweight image SR methods. The best is highlighted in \textcolor{red}{red} and the second is \underline{underlined}.} 
\label{tab:lightweight-sr}
\resizebox{0.95\textwidth}{!}{
\begin{threeparttable}
\begin{tabular}{l*{9}{c}}
\bottomrule[2pt]

\multirow{2}{*}{Method} & \multirow{2}{*}{Scale} & \multirow{2}{*}{\#Params} &   Set5      &   Set14     & BSD100    & Urban100    & Manga109   \\ \cmidrule(r){4-10}
& & &   PSNR/SSIM &   PSNR/SSIM & PSNR/SSIM   & PSNR/SSIM & PSNR/SSIM   \\ \midrule

CARN & \multirow{5}{*}{$\times2$}  & 1592K    & 37.76 / 0.9590       & 33.52 / 0.9166       & 32.09 / 0.8978       & 31.92 / 0.9256     & 38.36 / 0.9756   \\
IMDN & & 694K     & 38.00 / 0.9605       & 33.63 / 0.9177       & 32.19 / 0.8996       & \underline{32.17} / 0.9283     & 38.88 / 0.9774   \\
LAPAR-A & & 548K  & 38.01 / 0.9605       & 33.62 / 0.9183       & 32.19 / 0.8999       & 32.10 / 0.9283     & 38.67 / 0.9772   \\
SwinIR-T & & 878K  & \underline{38.14} / \underline{0.9611}       & \underline{33.86} / \underline{0.9206}       & \underline{32.31} / \underline{0.9012}       & \textcolor{red}{32.76} / \textcolor{red}{0.9340}     & \underline{39.12} / \underline{0.9783}   \\ \hdashline
\textbf{MMA-T} (Ours) & & 796K    & \textcolor{red}{38.17} / \textcolor{red}{0.9616}       & \textcolor{red}{33.96} / \textcolor{red}{0.9208}       & \textcolor{red}{32.34} / \textcolor{red}{0.9027}       & \textcolor{red}{32.76} / \underline{0.9339}     & \textcolor{red}{39.37} / \textcolor{red}{0.9790}   \\ \midrule[1pt]

CARN & \multirow{5}{*}{$\times3$}  & 1592K    & 34.29 / 0.9255       & 30.29 / 0.8407       & 29.06 / 0.8034       & 28.06 / 0.8493     & 33.50 / 0.9440   \\
IMDN & & 703K     & 34.36 / 0.9270       & 30.32 / 0.8417       & 29.09 / 0.8046       & 28.17 / 0.8519     & 33.61 / 0.9445   \\
LAPAR-A & & 544K  & 34.36 / 0.9267       & 30.34 / 0.8421       & \underline{29.11} / 0.8054       & 28.15 / 0.8523     & 33.51 / 0.9441   \\
SwinIR-T & & 886K  & \textcolor{red}{34.62} / 0.9289       & \textcolor{red}{30.54} / \textcolor{red}{0.8463}       & \textcolor{red}{29.20} / \underline{0.8082}       & \textcolor{red}{28.66} / \textcolor{red}{0.8624}     & \textcolor{red}{33.98} / \textcolor{red}{0.9478}   \\ \hdashline
\textbf{MMA-T} (Ours) & & 899K    & 34.57 / \textcolor{red}{0.9294}       & \underline{30.45} / \underline{0.8455}       & \textcolor{red}{29.20} / \textcolor{red}{0.8103}       & \underline{28.64} / \underline{0.8613}     & \underline{33.71} / \underline{0.9474}   \\ \midrule[1pt]

CARN & \multirow{5}{*}{$\times4$}  & 1592K    & 32.13 / 0.8937       & 28.60 / 0.7806       & 27.58 / 0.7349       & 26.07 / 0.7837     & 30.47 / 0.9084   \\
IMDN & & 715K     & 32.21 / 0.8948       & 28.58 / 0.7811       & 27.56 / 0.7353       & 26.04 / 0.7838     & 30.45 / 0.9075   \\
LAPAR-A & & 659K  & 32.15 / 0.8944       & 28.61 / 0.7818       & 27.61 / 0.7366       & 26.14 / 0.7871     & 30.42 / 0.9074   \\
SwinIR-T & & 897K  & \underline{32.44} / \underline{0.8976}       & \underline{28.77} / \underline{0.7858}       & \underline{27.69} / \underline{0.7406}       & \underline{26.47} / \underline{0.7980}     & \underline{30.92} / \underline{0.9151}   \\ \hdashline
\textbf{MMA-T} (Ours) & & 879K    & \textcolor{red}{32.50} / \textcolor{red}{0.8994}       & \textcolor{red}{28.80} / \textcolor{red}{0.7871}       & \textcolor{red}{27.72} / \textcolor{red}{0.7424}       & \textcolor{red}{26.54} / \textcolor{red}{0.7983}     & \textcolor{red}{31.13} / \textcolor{red}{0.9160}   \\ \bottomrule[2pt] 

\end{tabular}
\end{threeparttable}
}
\end{table*}

\section{Conclusion}

In this paper, we present an innovative SR model MMA, based on Vision Mamba (Vim).
MMA is composed of three parts: shallow feature extraction, deep feature extraction, and high-quality reconstruction modules.
Specifically, to better unleash the global representation ability of Vim, we identify three key points: 1) incorporating Vim into a MetaFormer-style block; 2) pre-training Vim-based models on a larger and broader dataset; 3) applying complementary CNN-based attention mechanism in a parallel structure.
Impartial and extensive experiments demonstrate that owing to a wider range of activated areas, MMA not only poses leading quantitative results but also restores vivid and authentic textures and details compared to existing image SR methods.
Furthermore, MMA also proves its versatility in lightweight SR applications. 
This exploration seeks to shed light on the expansive utility of SSMs within the domain of image processing, extending beyond SISR.
It is our aspiration that this work will serve as a catalyst for further innovative investigations in this evolving field, prompting a broader exploration of SSMs' application across various image processing challenges.
\bibliographystyle{splncs04}
\bibliography{main}
\end{document}